\documentclass[11pt]{article}
\usepackage{acl2016}
\usepackage{times}
\usepackage{latexsym}
\usepackage{amsmath,bm}
\usepackage{amsfonts}
\usepackage{mathrsfs}
\usepackage{amsbsy,amsthm,amssymb}

\usepackage{graphicx}
\usepackage{xcolor}
\usepackage{multirow}
\usepackage{paralist}
\usepackage{subfigure}

\usepackage{fancyhdr} 

\aclfinalcopy 

\title{Natural Language Inference by Tree-Based Convolution\\ and Heuristic Matching}

\author{Lili Mou\thanks{\ \ Equal contribution.\quad $^\dag$Corresponding authors.},\,$^{1}$ Rui Men,\hspace{-.11cm}$^{*\hspace{0.01cm}1}$ Ge Li,\!$^{\dag1}$ Yan Xu,$^{1}$ Lu Zhang,$^{1}$ Rui Yan,$^2$ Zhi Jin$^{\dag1}$\\
\normalsize $^{1}$Key Laboratory of High Confidence Software Technologies (Peking University),\\\normalsize
Ministry of Education, China;\ \ Software Institute, Peking University, China\\\normalsize
{\tt \{doublepower.mou,menruimr\}@gmail.com}\\ \normalsize
{\tt \{lige,xuyan14,zhanglu,zhijin\}@sei.pku.edu.cn}\\\normalsize
$^{2}$Baidu Inc., China\quad {\tt yanrui02@baidu.com}
}

\date{}

\begin{document}

\renewcommand{\headrulewidth}{0pt}
\cfoot{Accpeted by ACL'16 as a short paper}
\thispagestyle{fancy}

\maketitle

\begin{abstract}
In this paper, we propose the TBCNN-pair model to recognize entailment and contradiction between two sentences. In our model, a tree-based convolutional neural network (TBCNN) captures sentence-level semantics; then heuristic matching layers like concatenation, element-wise product/difference combine the information in individual sentences. Experimental results show that our model outperforms existing sentence encoding-based approaches by a large margin.
\end{abstract}

\section{Introduction}

Recognizing entailment and contradiction between two sentences (called a \textit{premise} and a \textit{hypothesis}) is known as \textit{natural language inference} (NLI) in \newcite{inference}. Provided with a premise sentence, the task is to judge whether the hypothesis can be inferred ({\tt entailment}), or the hypothesis cannot be true ({\tt contradiction}). Several examples are illustrated in Table~\ref{tab:example}.

NLI is in the core of natural language understanding and has wide applications in NLP, e.g., question answering \cite{QA} and automatic summarization \cite{summarization,summarization2,summarization3}. Moreover, NLI is also related to other tasks of sentence pair modeling, including  paraphrase detection \cite{CNN:NIPS}, relation recognition of discourse units \cite{DRR-aaai}, etc.

Traditional approaches to NLI mainly fall into two groups: feature-rich models and formal reasoning methods. Feature-based approaches typically leverage machine learning models, but require intensive human engineering to represent lexical and syntactic information in two sentences \cite{feature1,feature2}. Formal reasoning, on the other hand, converts a sentence into a formal logical representation and uses interpreters to search for a proof. However, such approaches are limited in terms of scope and accuracy \cite{formal}. 

The renewed prosperity of neural networks has made significant achievements in various NLP applications, including individual sentence modeling \cite{sentenceCNN,sentenceTBCNN} as well as sentence matching \cite{CNN:NIPS,CNN:NAACL}. A typical neural architecture to model sentence pairs is the ``Siamese'' structure \cite{siamese}, which involves an underlying sentence model and a matching layer to determine the relationship between two sentences.  Prevailing sentence models include convolutional networks \cite{sentenceCNN} and recurrent/recursive networks \cite{recursive}. Although they have achieved high performance, they may either fail to fully make use of the syntactical information in sentences or be difficult to train due to the long propagation path. Recently, we propose a novel tree-based convolutional neural network (TBCNN) to alleviate the aforementioned problems and have achieved higher performance in two sentence classification tasks \cite{sentenceTBCNN}. However, it is less clear whether TBCNN can be harnessed to model sentence pairs for implicit logical inference, as is in the NLI task.

\begin{table}[!t]
\vspace{-.2cm}
\centering
\resizebox{.48\textwidth}{!}{
\begin{tabular}{l|l|l}
\hline
\!\!\textbf{Premise} & Two men on bicycles competing in a race.\\
\hline
                    & People are riding bikes. &\! {\tt E}\!\!\\
\!\!\textbf{Hypothesis} & Men are riding bicycles on the streets. &\! {\tt C}\!\!\\
                    & A few people are catching fish. &\! {\tt N}\!\!\\
\hline
\end{tabular}
}
\vspace{-.4cm}
\caption{Examples of relations between a premise and a hypothesis: {\tt E}ntailment, {\tt C}ontradiction, and {\tt N}eutral (irrelevant).}
\vspace{-.3cm}
\label{tab:example}
\end{table}

In this paper, we propose the TBCNN-pair neural model to recognize entailment and contradiction between two sentences. We leverage our newly proposed TBCNN model to capture structural information in sentences, which is important to NLI. For example, the phrase ``riding bicycles on the streets'' in Table~\ref{tab:example} can be well recognized by TBCNN via the dependency relations {\tt dobj(riding,bicycles)} and {\tt prep\_on(riding,street)}. As we can see, TBCNN is more robust than sequential convolution in terms of word order distortion, which may be introduced by determinators, modifiers, etc. A pooling layer then aggregates information along the tree, serving as a way of semantic compositonality. Finally, two sentences' information is combined by several heuristic matching layers, including concatenation, element-wise product and difference; they are effective in capturing relationships between two sentences, but remain low complexity.

To sum up, the main contributions of this paper are two-fold: (1) We are the first to introduce tree-based convolution to sentence pair modeling tasks like NLI; (2) Leveraging additional heuristics further improves the accuracy while remaining low complexity, outperforming existing sentence encoding-based approaches to a large extent, including feature-rich methods and long short term memory (LSTM)-based recurrent networks.\footnote{
Code is released on:\\
{\color{white}}\quad\quad {https://sites.google.com/site/tbcnninference/}\\
}
 

\vspace{-.1cm}
\section{Related Work}\label{sec:Related}

\vspace{-.1cm}
Entailment recognition can be viewed as a task of sentence pair modeling. Most neural networks in this field involve a sentence-level model, followed by one or a few matching layers. They are sometimes called ``Siamese'' architectures \cite{siamese}.

\newcite{CNN:NIPS} and \newcite{CNN:NAACL} apply convolutional neural networks (CNNs) as the individual sentence model, where a set of feature detectors over successive words are designed to extract local features. \newcite{LSTM:AAAI} build sentence pair models upon recurrent neural networks (RNNs) to iteratively integrate information along a sentence. \newcite{recurparaphrase} dynamically construct tree structures (analogous to parse trees) by recursive autoencoders to detect paraphrase between two sentences. As shown, inherent structural information in sentences is oftentimes important to natural language understanding.


The simplest approach to match two sentences, perhaps, is to concatenate their vector representations \cite[Arc-I]{DRR,CNN:NIPS}. Concatenation is also applied in our previous work of matching the subject and object in relation classification \cite{relation,relation2}. \newcite{CNN:EMNLP} apply additional heuristics, namely Euclidean distance, cosine measure, and element-wise absolute difference.
The above methods operate on a fixed-size vector representation of a sentence, categorized as \textit{sentence encoding}-based approaches. Thus the matching complexity is $\mathcal{O}(1)$, i.e., independent of the sentence length. Word-by-word similarity matrices are introduced to enhance interaction. To obtain the similarity matrix, \newcite{CNN:NIPS} (Arc-II) concatenate two words' vectors (after convolution), \newcite{recurparaphrase} compute Euclidean distance, and \newcite{LSTM:AAAI} apply tensor product. In this way, the complexity is of $\mathcal{O}(n^2)$, where $n$ is the length of a sentence; hence similarity matrices are difficult to scale and less efficient for large datasets.

Recently, \newcite{attention} introduce several context-aware methods for sentence matching. They report that RNNs over a single chain of two sentences are more informative than separate RNNs; a static attention over the first sentence is also useful when modeling the second one. Such context-awareness interweaves the sentence modeling and matching steps. In some scenarios like sentence pair re-ranking \cite{sigir}, it is not feasible to pre-calculate the vector representations of sentences, so the matching complexity is of $\mathcal{O}(n)$. \newcite{attention} further develop a word-by-word attention mechanism and obtain a higher accuracy with a complexity order of $\mathcal{O}(n^2)$.
 
\vspace{-.1cm}
\section{Our Approach}\label{sec:Approach}
\vspace{-.1cm}

\begin{figure}[!t]
\centering
\includegraphics[width=.49\textwidth]{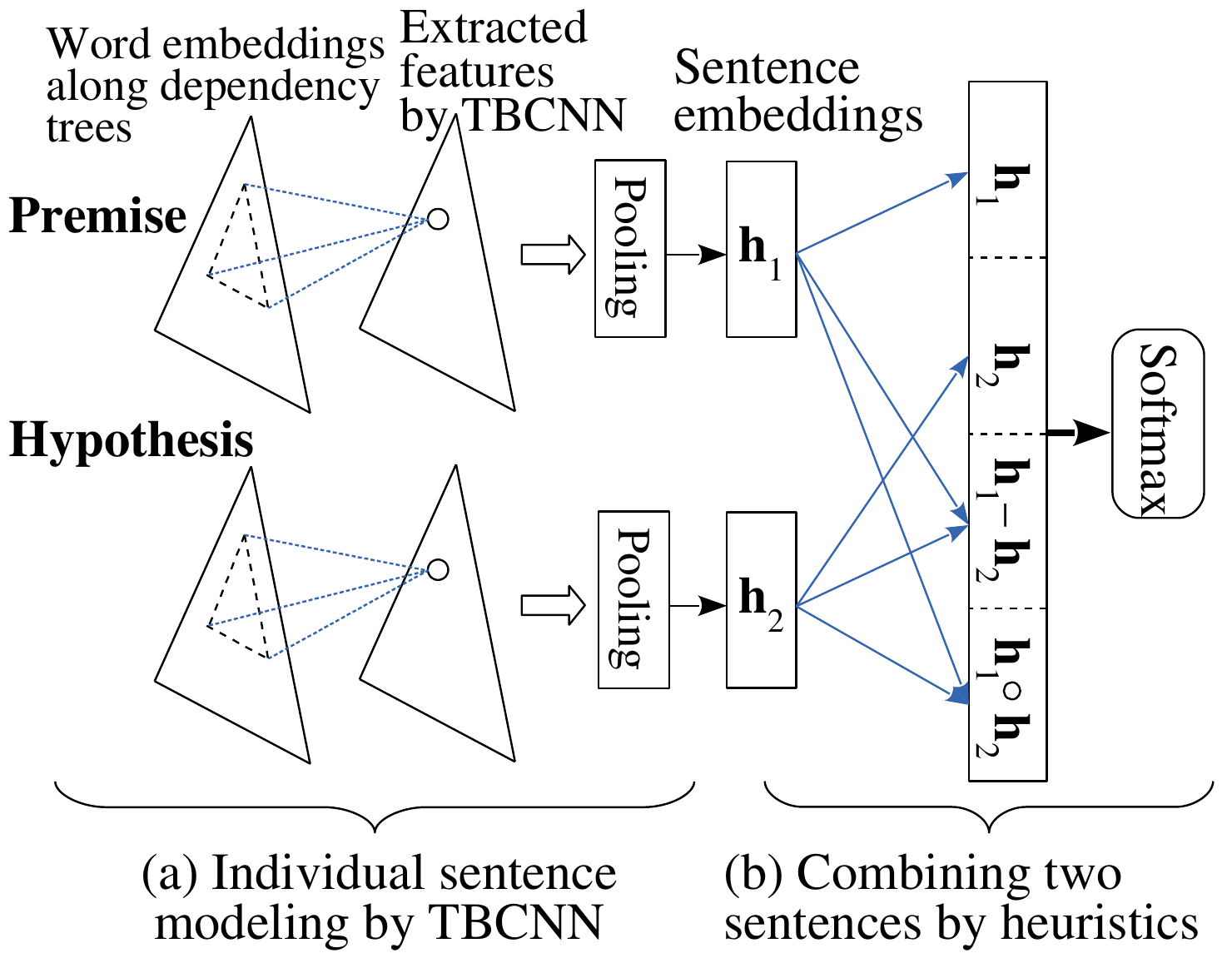}
\vspace{-.8cm plus 0cm minus 0cm}
\caption{TBCNN-pair model. (a) Individual sentence modeling via tree-based convolution. (b) Sentence pair modeling with heuristics, after which a softmax layer is applied for output.}\label{fig:model}

\end{figure}
We follow the ``Siamese'' architecture (like most work in Section~\ref{sec:Related}) and adopt a two-step strategy to classify the relation between two sentences. Concretely, our model comprises two parts:

\begin{compactitem}[\ \ $\bullet$]
\item A tree-based convolutional neural network models each individual sentence (Figure~\ref{fig:model}a). Notice that, the two sentences, premise and hypothesis, share a same TBCNN model (with same parameters), because this part aims to capture general semantics of sentences.

\item A matching layer combines two sentences' information by heuristics (Figure~\ref{fig:model}b). After individual sentence models, we design a sentence matching layer to aggregate information. We use simple heuristics, including concatenation, element-wise product and difference, which are effective and efficient.
\end{compactitem}

Finally, we add a softmax layer for output. The training objective is cross-entropy loss, and we adopt mini-batch stochastic gradient descent, computed by back-propagation.

\vspace{-.1cm}
\subsection{Tree-Based Convolution}

\vspace{-.1cm}
The tree-based convolutoinal neural network (TBCNN) is first proposed in our previous work \cite{programTBCNN}\footnote{Preprinted on arXiv on September 2014\\ {\color{white}} \ \ \ \ \ \ \  \ (http://arxiv.org/abs/1409.5718v1)
} to classify program source code; later, we further propose TBCNN variants to model sentences \cite{sentenceTBCNN}. 
This subsection details the tree-based convolution process.

The basic idea of TBCNN is to design a set of subtree feature detectors sliding over the parse tree of a sentence; either a constituency tree or a dependency tree applies. In this paper, we prefer the dependency tree-based convolution for its efficiency and compact expressiveness.

Concretely, a sentence is first converted to a dependency parse tree.\footnote{Parsed by the Stanford parser\\ {\color{white}} \ \ \ \ \ \ \  \ (http://nlp.stanford.edu/software/lex-parser.shtml)
} Each node in the dependency tree corresponds to a word in the sentence; an edge $a\!\!\rightarrow\!\!b$ indicates $a$ is governed by $b$. Edges are labeled with grammatical relations (e.g., {\tt nsubj}) between the parent node and its children \cite{dependency}. Words are represented by pretrained vector representations, also known as \textit{word embeddings} \cite{word2vec}.
 
Now, we consider a set of two-layer subtree feature detectors sliding over the dependency tree. At a position where the parent node is $p$ with child nodes $c_1, \cdots, c_n$, the output of the feature detector, $\bm y$, is 

\vspace{-1.1cm}
$$\bm y=f\left(W_p \bm p + \sum_{i=1}^nW_{r[c_i]}\bm c_i+\bm b\right)$$

\vspace{-.2cm}
Let us assume word embeddings ($\bm p$ and $\bm c_i$) are of $n_e$ dimensions; that the convolutional layer $\bm y$ is $n_c$-dimensional. $W\in\mathbb{R}^{n_c\times n_e}$ is the weight matrix; $\bm b\in\mathbb{R}^{n_c}$ is the bias vector. $r[c_i]$ denotes the dependency relation between $p$ and $c_i$. $f$ is the non-linear activation function, and we apply {\tt ReLU} in our experiments.

 After tree-based convolution, we obtain a set of feature maps, which are one-one corresponding to original words in the sentence. Therefore, they may vary in size and length. A dynamic pooling layer is applied to aggregate information along different parts of the tree, serving as a way of \textit{semantic compositionality} \cite{CNN:NIPS}. We use the $\max$ pooling operation, which takes the maximum value in each dimension. 

Then we add a fully-connected hidden layer to further mix the information in a sentence. The obtained vector representation of a sentence is denoted as $\bm h$ (also called a \textit{sentence embedding}). Notice that the same tree-based convolution applies to both the  premise and hypothesis.

Tree-based convolution along with pooling enables structural features to reach the output layer with short propagation paths, as opposed to the recursive network \cite{recursive}, which is also structure-sensitive but may suffer from the problem of long propagation path. By contrast, TBCNN is effective and efficient in learning such structural information \cite{sentenceTBCNN}.

\subsection{Matching Heuristics}\label{ss:matching}

\vspace{-.1cm}
In this part, we introduce how vector representations of individual sentences are combined to capture the relation between the premise and hypothesis. As the dataset is large, we prefer $\mathcal{O}(1)$ matching operations because of efficiency concerns.  Concretely, we have three matching heuristics:
\begin{compactitem}
\item Concatenation of the two sentence vectors,
\item Element-wise product, and
\item Element-wise difference.
\end{compactitem}
The first heuristic follows the most standard procedure of the ``Siamese'' architectures, while the latter two are certain measures of ``similarity'' or ``closeness.''
These matching layers are further concatenated (Figure~\ref{fig:model}b), given by 

\vspace{-.3cm plus 0cm minus 0cm}
$$\bm m=[\bm h_1; \bm h_2; \bm h_1-\bm h_2;\bm h_1\circ\bm h_2]$$
\vspace{-.5cm plus 0cm minus 0cm}

\noindent where $\bm h_1\in\mathbb{R}^{n_c}$ and $\bm h_2\in\mathbb{R}^{n_c}$ are the sentence vectors of the premise and hypothesis, respectively; ``$\circ$'' denotes element-wise product; semicolons refer to column vector concatenation. $\bm m\in\mathbb{R}^{4n_c}$ is the output of the matching layer.

We would like to point out that, with subsequent linear transformation, element-wise difference is a special case of concatenation. 
If we assume the subsequent transformation takes the form of $W[\bm h_1\ \bm h_2]^\top$,
where $W\!\!=\!\![W_1\ W_2]$ is the weights for concatenated sentence representations, then element-wise difference can be viewed as such that $W_0(\bm h_1-\bm h_2)=[W_0\ -\!W_0][\bm h_1\ \bm h_2]^\top$. ($W_0$ is the weights corresponding to element-wise difference.)
Thus, our third heuristic can be absorbed into the first one in terms of model capacity. However, as will be shown in the experiment, explicitly specifying this heuristic significantly improves the performance, indicating that optimization differs, despite the same model capacity.
Moreover, word embedding studies show that linear offset of vectors can capture
relationships between two words \cite{offset}, but it has not been exploited in sentence-pair relation recognition. Although element-wise distance is used to detect paraphrase in \newcite{CNN:EMNLP}, it mainly reflects ``similarity'' information. Our study verifies that vector offset is useful in capturing generic sentence relationships, akin to the word analogy task.

\vspace{-.1cm plus 0cm minus 0cm}
\section{Evaluation}\label{sec:Result}
\vspace{-.2cm plus 0cm minus 0cm}
\subsection{Dataset}\label{ss:dataset}
\vspace{-.1cm plus 0cm minus 0cm}

To evaluate our TBCNN-pair model, we used the newly published Stanford Natural Language Inference (SNLI) dataset \cite{NLI}.\footnote{http://nlp.stanford.edu/projects/snli/
} The dataset is constructed by crowdsourced efforts, each sentence written by humans. 
Moreover, the SNLI dataset is magnitudes of larger than previous resources, and hence is particularly suitable for comparing neural models.
The target labels comprise three classes: {\tt Entailment}, {\tt Contradiction}, and {\tt Neutral} (two irrelevant sentences). 
We applied the standard train/validation/test split, contraining 550k, 10k, and 10k samples, respectively. Figure~\ref{tab:statistics} presents additional dataset statistics, especially those relevant to dependency parse trees.\footnote{We applied \textit{collapsed} dependency trees, where prepositions and conjunctions are annotated on the dependency relations, but these auxiliary words themselves are removed.}

\begin{table}[!t]
\centering
\begin{tabular}{|ccc|}
\hline
\textbf{Statistics} & \textbf{Mean} &\textbf{Std}\\
\hline\hline
\# nodes   & 8.59 & 4.14\\
Max depth & 3.93 & 1.13\\
Avg leaf depth & 3.13 & 0.65\\
Avg node depth & 2.60 & 0.54\\
\hline
\end{tabular}
\caption{Statistics of the Stanford Natural Language Inference dataset where each sentence is parsed into a dependency parse tree.}\label{tab:statistics}
\end{table}
\begin{figure}[!t]
\centering
\includegraphics[width=.35\textwidth]{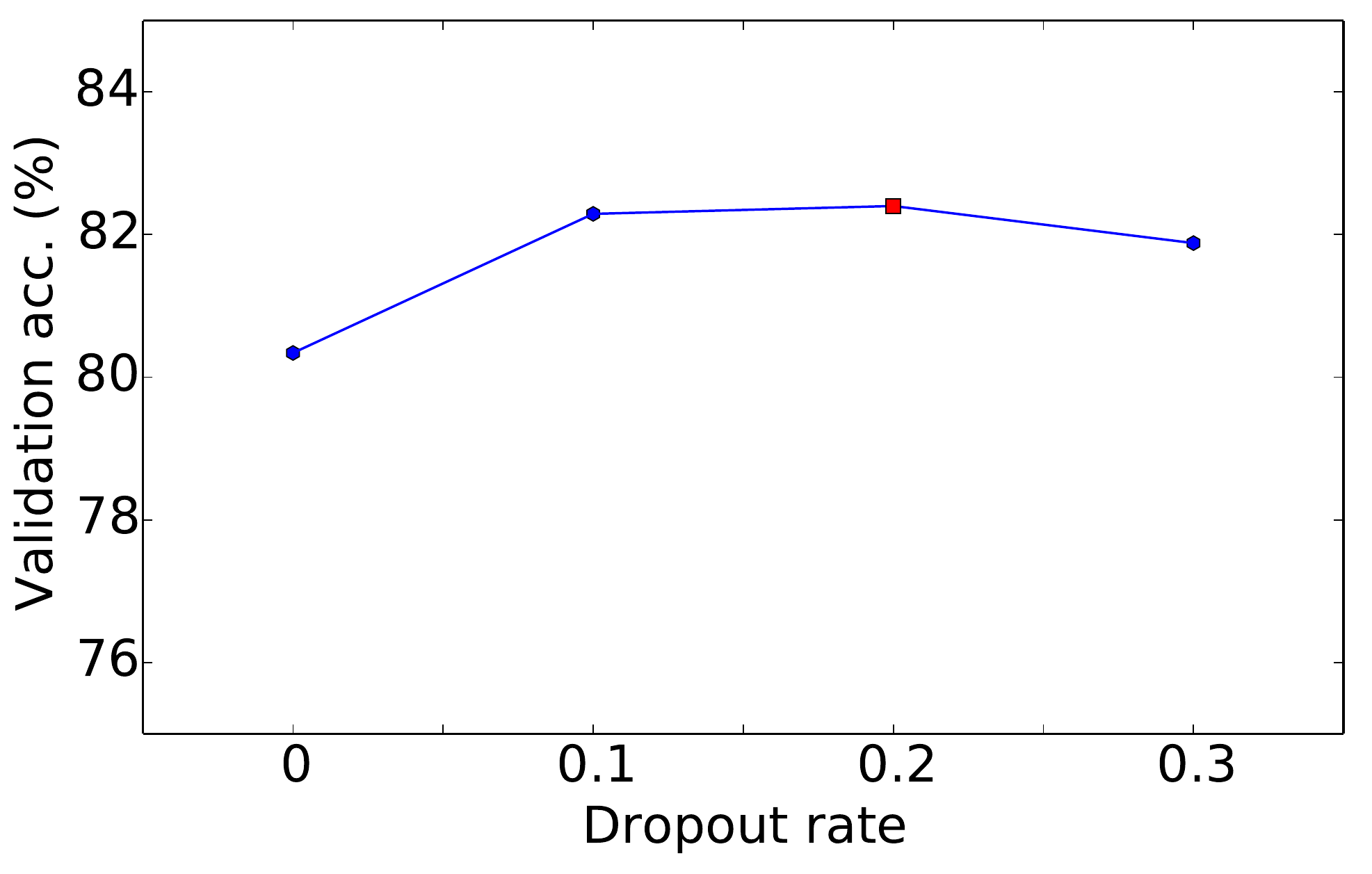}
\vspace{-.3cm}
\caption{Validation accuracy versus dropout rate (full TBCNN-pair model).}\label{fig:dropout}
\end{figure}

\subsection{Hyperparameter Settings}\label{ss:hyperparameters}

All our neural layers, including embeddings, were set to 300 dimensions. The model is mostly robust when the dimension is large, e.g., several hundred \cite{unified}. Word embeddings were pretrained ourselves by {\tt word2vec} on the English Wikipedia corpus and fined tuned during training as a part of model parameters.  We applied $\ell_2$ penalty of $3\times10^{-4}$; dropout was chosen by validation with a granularity of 0.1 (Figure~\ref{fig:dropout}). We see that a large dropout rate ($\ge$ 0.3) hurts the performance (and also makes training slow) for such a large dataset as opposed to small datasets in other tasks \cite{regularization}. Initial learning rate was set to 1, and a power decay was applied. We used stochastic gradient descent with a batch size of 50.

\subsection{Performance}\label{ss:result}
\begin{table}[!t]
\centering
\resizebox{.48\textwidth}{!}{
\begin{tabular}{|lr|c|}
\hline
\multirow{2}*{\!\!\textbf{Model}} 
&\!\!\!\!\!\!\!\!\!\!\!\!\!\!\!\!\textbf{Test acc.}\!\!\!&\!\! \textbf{Matching}\!\!\\
               &  \textbf{(\%)}\!\!               & \!\!\!\!\textbf{complexity}\!\!\!\! \\
\hline\hline
\!\!Unlexicalized features$^b$ & 50.4\!\! & \multirow{10}*{$\mathcal{O}(1)$}\\
\!\!Lexicalized features$^b$   & 78.2\!\!&\\
\cline{1-2}
\!\!Vector sum  $+$ MLP$^b$     & 75.3\!\!&\\  
\!\!Vanilla RNN $+$ MLP$^b$   & 72.2\!\!&\\
\!\!LSTM    RNN $+$ MLP$^b$     & 77.6\!\!& \\
\!\!CNN $+$ cat & 77.0\!\!&\\
\!\!GRU w/ skip-thought pretraining$^v$\!\!\!\! & 81.4\!\!&\\
\cline{1-2}
\!\!TBCNN-pair $+$ cat            &  79.3\!\!&\\
\!\!TBCNN-pair $+$ cat,$\circ$,-  &  \textbf{82.1}\!\!&\\
\hline
\!\!Single-chain LSTM RNNs$^r$ & 81.4\!\!&  \multirow{2}*{$\mathcal{O}(n)$}\\
$+$ static attention$^r$ &   \textbf{82.4}\!\!&\\
\hline
\!\!LSTM $+$ word-by-word attention$^r$\!\!\!\! & \textbf{83.5}\!\! &$\mathcal{O}(n^2)$\\
\hline
\end{tabular}
}
\caption{Accuracy of the TBCNN-pair model in comparison with previous results ($^b$Bowman et al., 2015; $^v$Vendrov et al., 2015; $^r$Rockt{\"a}schel et al., 2015). ``cat'' refers to concatenation; ``-'' and ``$\circ$'' denote element-wise difference and product, resp.}\label{tab:result}
\end{table}

\begin{table}[!t]
\centering
\resizebox{!}{!}{
\begin{tabular}{|l|cc|}
\hline
\textbf{Model Variant} & \textbf{Valid Acc.} & \textbf{Test Acc.}\\
\hline\hline
TBCNN$+\circ$         & 73.8   & 72.5\\
TBCNN$+$-             & 79.9   & 79.3\\
TBCNN$+$cat           & 80.8   & 79.3 \\
\hline
TBCNN$+$cat,$\circ$   & 81.6   & 80.7  \\
TBCNN$+$cat,-         & 81.7   & 81.6 \\
TBCNN$+$cat,$\circ$,- & \textbf{82.4}   & \textbf{82.1}\\ 
\hline
\end{tabular}
}
\caption{Validation and test accuracies of TBCNN-pair variants (in percentage).}\label{tab:variant}
\end{table}
Table~\ref{tab:result} compares our model with previous results. As seen, the TBCNN sentence pair model, followed by simple concatenation alone, outperforms existing sentence encoding-based approaches (without pretraining), including a feature-rich method using 6 groups of human-engineered features, long short term memory (LSTM)-based RNNs, and traditional CNNs. This verifies the rationale for using tree-based convolution as the sentence-level neural model for NLI.

Table~\ref{tab:variant} compares different heuristics of matching. We first analyze each heuristic separately: using element-wise product alone is significantly worse than concatenation or element-wise difference; the latter two are comparable to each other. 

Combining different matching heuristics improves the result: the TBCNN-pair model with concatenation, element-wise product and difference yields the highest performance of 82.1\%. As analyzed in Section~\ref{ss:matching}, the element-wise difference matching layer does not add to model complexity and can be absorbed as a special case into simple concatenation. However, explicitly using such heuristic yields an accuracy boost of 1--2\%. Further applying element-wise product improves the accuracy by another 0.5\%.

The full TBCNN-pair model outperforms all existing sentence encoding-based approaches, including a 1024d gated recurrent unit (GRU)-based RNN with ``skip-thought'' pretraining \cite{skipthought}. The results obtained by our model are also comparable to several attention-based LSTMs, which are more computationally intensive than ours in terms of complexity order. 

\subsection{Complexity Concerns}

For most sentence models including TBCNN, the overall complexity is at least $\mathcal{O}(n)$. However, an efficient matching approach is still important, especially to \textit{retrieval-and-reranking} systems \cite{sigir,ijcai}. For example, in a retrieval-based question-answering or conversation system, we can largely reduce response time by performing sentence matching based on precomputed candidates' embeddings. By contrast, context-aware matching approaches as described in Section~\ref{sec:Related} involve processing each candidate given a new user-issued query, which is time-consuming in terms of most industrial products. 

In our experiments, the matching part (Figure~\ref{fig:model}b) counts 1.71\% of the total time during prediction (single-CPU, \texttt{C++} implementation), showing the potential applications of our approach in efficient retrieval of semantically related sentences.

\section{Conclusion}\label{sec:Conclusion}

In this paper, we proposed the TBCNN-pair model for natural language inference. Our model relies on the tree-based convolutional neural network (TBCNN) to capture sentence-level semantics; then two sentences' information is combined by several heuristics including concatenation, element-wise product and difference. Experimental results on a large dataset show a high performance of our TBCNN-pair model while remaining a low complexity order.

\section*{Acknowledgments}

We thank all anonymous reviewers for their constructive comments, especially those on complexity issues. We also thank Sam Bowman, Edward Grefenstette, and Tim Rockt\"aschel for their discussion. This research was supported by the National Basic Research Program of China (the 973 Program) under Grant No.~2015CB352201 and the National Natural Science Foundation of China under Grant Nos.~61232015, 61421091, and 61502014.

\newpage
\bibliographystyle{acl2016}
\bibliography{inference}

\end{document}